\documentclass[10pt,twocolumn,letterpaper]{article}

\usepackage{amsmath}
\usepackage{amssymb}
\usepackage{cvpr}
\usepackage{epsfig}
\usepackage{graphicx}
\usepackage{times}

\usepackage{subcaption}
\usepackage{tabularx}
\usepackage[title]{appendix}
\usepackage{hyperref}
\hypersetup{
    colorlinks=true,
    linkcolor=blue,
    filecolor=magenta,      
    urlcolor=cyan,
    pdftitle={Sharelatex Example},
    pdfpagemode=FullScreen,
}

\cvprfinalcopy 

\def\cvprPaperID{****} 

\newcommand{\Csh}{{\settoheight{\dimen0}{C}C\kern-.05em \resizebox{!}{\dimen0}{\raisebox{\depth}{\textbf{\#}}}}}

\begin{document}

\title{Unity Perception: Generate Synthetic Data for Computer Vision}

\author{Steve Borkman, Adam Crespi, Saurav Dhakad, Sujoy Ganguly, Jonathan Hogins, \\ 
You-Cyuan Jhang, Mohsen Kamalzadeh, Bowen Li, Steven Leal, Pete Parisi, Cesar Romero, \\
Wesley Smith, Alex Thaman, Samuel Warren, Nupur Yadav  \\
\textbf{Unity Technologies}\\
{\tt\small computer-vision@unity.com}
}

\maketitle


\begin{abstract}

We introduce the Unity Perception package which aims to simplify and accelerate the process of generating synthetic datasets for computer vision tasks by offering an easy-to-use and highly customizable toolset. This open-source package extends the Unity Editor and engine components to generate perfectly annotated examples for several common computer vision tasks. Additionally, it offers an extensible Randomization framework that lets the user quickly construct and configure randomized simulation parameters in order to introduce variation into the generated datasets. We provide an overview of the provided tools and how they work, and demonstrate the value of the generated synthetic datasets by training a 2D object detection model. The model trained with mostly synthetic data outperforms the model trained using only real data. 

\end{abstract}

\section{Introduction}

Over the past decade, computer vision has evolved from heuristics to a set of deep-learning-based models that learn from large amounts of labeled data. These models have made significant progress in what are now considered more basic tasks like image classification, and the computer vision community has shown interest in more complex tasks such as object detection\cite{redmon_you_2016, liu_ssd_2016, lin_focal_2017, ren_faster_2017, he_mask_2017, tan_efficientdet_2020}, semantic segmentation\cite{long_fully_2015, chen_deeplab_2017, ronneberger_u-net_2015, chen_rethinking_2017}, and instance segmentation\cite{he_mask_2017, kuo_shapemask_2019, bolya_yolact_2019}. These more complex tasks require increasingly complex models, datasets, and labels. However, these large and complex datasets come with challenges related to cost, bias, and privacy.

As the focus shifts to more complex tasks, the cost of annotating each example increases from labeling frames to labeling objects and even pixels in the image. Furthermore, as human annotators label these examples, their workflows and tools become more complex as well. This, in turn, creates a need to review or audit annotations, leading to additional costs for each labeled example.
Furthermore, as tasks become complex and the range of possible variations to account for expands, the requirements of data collection become more challenging. Some scenarios may rarely occur in the real world, yet correctly handling these events is crucial. For example, misplaced obstacles on the road need to be detected by autonomous vehicles. Such rare events further increase the cost of collecting data and not accounting for them can affect the model's performance, leading to uncertainty about how the model will perform on these rare events.
Privacy concerns surrounding machine learning models have also become increasingly important, further complicating data collection. Regulations such as the EU General Data Protection Regulation (GDPR)\cite{voigt_eu_2017} and the California Consumer Privacy Act (CCPA)\cite{bukaty_california_2019} enhance privacy rights and restrict the use of consumer data to train machine learning models.

By creating virtual environments, we can generate training examples in a controllable and customizable manner and avoid the challenges of collecting and annotating data in the real world. A rendering engine requires compute-time rather than human-time to generate examples. It has perfect information about the scenes it renders, making it possible to bypass the time and cost of human annotations and reviews. A rendering engine also makes it possible to generate rare examples, allowing control on the distribution of the training dataset. Moreover, such a process does not rely on any individual's private data by design.

\begin{figure}
    \centering
    \includegraphics[width=0.9\linewidth]{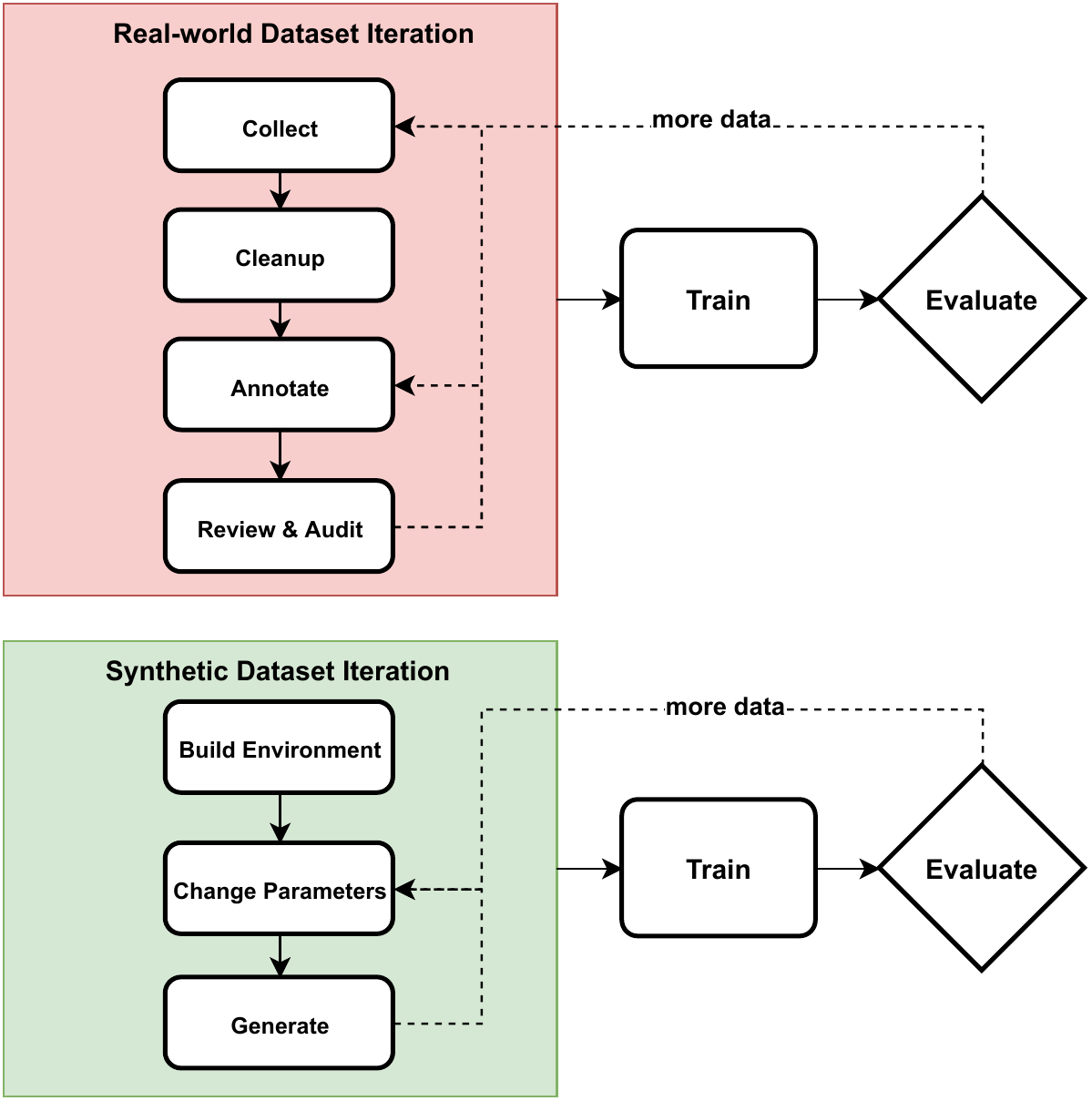}
    \caption{Comparison of the dataset iteration process with real-world and synthetic datasets. \textbf{Top:} At a minimum, a real-world dataset requires the collect, cleanup, annotate, and finally review, and audit steps. These steps involve costly and time-consuming human effort. \textbf{Bottom:} A synthetic dataset requires an environment built with 3D assets, setting or altering randomization parameters, and running the environment to generate new data. The datasets come with accurate annotations and are automatically validated, eliminating most of the time-consuming steps.}
    \label{fig:dataset-iteration}
\end{figure}

An additional benefit of using synthetic environments to generate labeled data is that the environments are reusable. Once we create the environment with all its visual assets, we can introduce variation into the environment at simulation time by changing randomization parameters on the fly. This allows for faster iterations on the generated datasets and the computer vision model. Figure \ref{fig:dataset-iteration} demonstrates how manually annotated datasets differ from synthetically generated ones in terms of workflows and iteration. By eliminating the need to review and audit datasets and avoiding time-consuming data collection and clean-up steps, synthetic data open the door to more rapid iteration.

In this work, we introduce the Unity Perception package\footnote{\url{https://github.com/Unity-Technologies/com.unity.perception}}, which offers a variety of convenient and customizable tools that help speed up and simplify the process of generating labeled synthetic datasets for computer vision problems. This open-source package extends the Unity editor and Unity's world-class rendering engine\cite{unity_unity_2021} with the capability to generate perfectly annotated examples for several common computer vision tasks (Section \ref{sec:labeling}). As described in Section \ref{sec:randomizer}, the package comes with an extensible randomization framework for setting up and configuring randomized parameters that can introduce variation into the generated datasets (Section \ref{sec:randomizer}). We also provide a companion python package to help consume and parse the generated datasets (Section \ref{sec:dataset-insights}).


\section{Related Work}

The computer vision community has invested great resources to create datasets such as PASCAL VOC\cite{everingham_pascal_2010}, NYU-Depth V2\cite{silberman_indoor_2012}, MS COCO\cite{lin_microsoft_2014}, and SUN RGB-D\cite{song_sun_2015}. 
While these have contributed to a significant boost in research on complex tasks such as semantic segmentation of indoor scenes\cite{chen_deeplab_2017, he_deep_2016, badrinarayanan_segnet_2017}, they cannot cover all the scenarios researchers are interested in, or provide a path for others to create new datasets. Some researchers have identified the potential for synthetic data and have achieved great results in specific tasks and domains. Examples of this include object detection of groceries\cite{hinterstoisser_annotation_2019}, controlling robotic arms to move blocks\cite{tobin_domain_2018}, and fine-grained manipulation of a Rubik's cube\cite{openai_solving_2019}; however, these simulation environments are often not publicly available, which makes the research difficult to reproduce and impossible to extend.

Others have identified this challenge and created simulation environments such as SYNTHIA\cite{ros_synthia_2016}, CARLA\cite{dosovitskiy_carla_2017}, AI2-THOR\cite{kolve_ai2-thor_2019}, Habitat\cite{savva_habitat_2019}, and iGibson\cite{shen_igibson_2020}. These simulators increase the number of examples researchers can use to train their computer vision models and facilitate reproducibility; however, they do not provide a general-purpose platform as they couple the simulator with specific domains and tasks. NVIDIA Isaac Sim\cite{noauthor_nvidia_2019} takes these efforts a step further and provides a platform built to enable a wide range of robotics simulations, instead of a specific domain or task. BlenderProc\cite{denninger_blenderproc_2019} and NVISII\cite{morrical_nvisii_2021} share similar goals of providing an API to generate training examples for a few computer vision tasks, but don't focus on physics or simulation.

The Unity Perception package builds on top of the Unity Editor to be a building block of such robotics and computer vision research projects. The first project to use this package was SynthDet, which is and end-to-end solution for detecting grocery objects using synthetic datasets and analyzing model performance under various combinations of real and synthetic data. This project will be discussed in section \ref{sec:synthdet}.

\begin{table*}[t]
    \centering
    \begin{tabular} { c|cccc } 
        &NVIDIA Isaac Sim&NVISII&BlenderProc& Unity Perception \\
           \hline
        Semantic segmentation&yes&yes&yes&yes \\
        Instance segmentation&yes&yes&yes&yes \\
        2D bounding-box&yes&yes&yes&yes \\
        3D bounding-box&yes&yes&yes&yes \\
        Depth&yes&yes&yes&no \\
        Keypoints&no&yes&no&yes \\
        Normals&--&yes&yes&no \\
        Optical flow&--&yes&no&no \\
        \hline
    \end{tabular}
    \caption{A comparison of different platforms and common computer vision tasks they support.}
    \label{tab:tasks}
\end{table*}

\begin{table*}[t]
    \centering
    \begin{tabular} { c|cccc } 
        &NVIDIA Isaac&NVISII&BlenderProc& Unity Perception \\
        \hline
        Content&--&--&--&asset store \\
        Domain Randomization&yes&--&no&yes\\
        Behavior simulation&yes&no&no&yes \\
        \hline
        Physics&yes&no&yes&yes \\
        Robotics&yes&--&no&yes \\
        Managed cloud scaling&no&no&no&yes \\
        Speed&++&++&+&++ \\
        Rendering&RT,R&RT&RT&RT(Windows only),R \\
        Language&C++, python&python&python&C\#
    \end{tabular}
    \caption{Summary of features provided by Unity Perception and other comparable platforms. '--' means unclear or partial support. RT means ray tracing and R means rasterization.}
    \label{tab:features}
\end{table*}

\section{The Unity Perception Package}

\begin{figure}[!h]
\hspace*{-1mm}
\begin{tabular}{cc}
 \includegraphics[width=38mm]{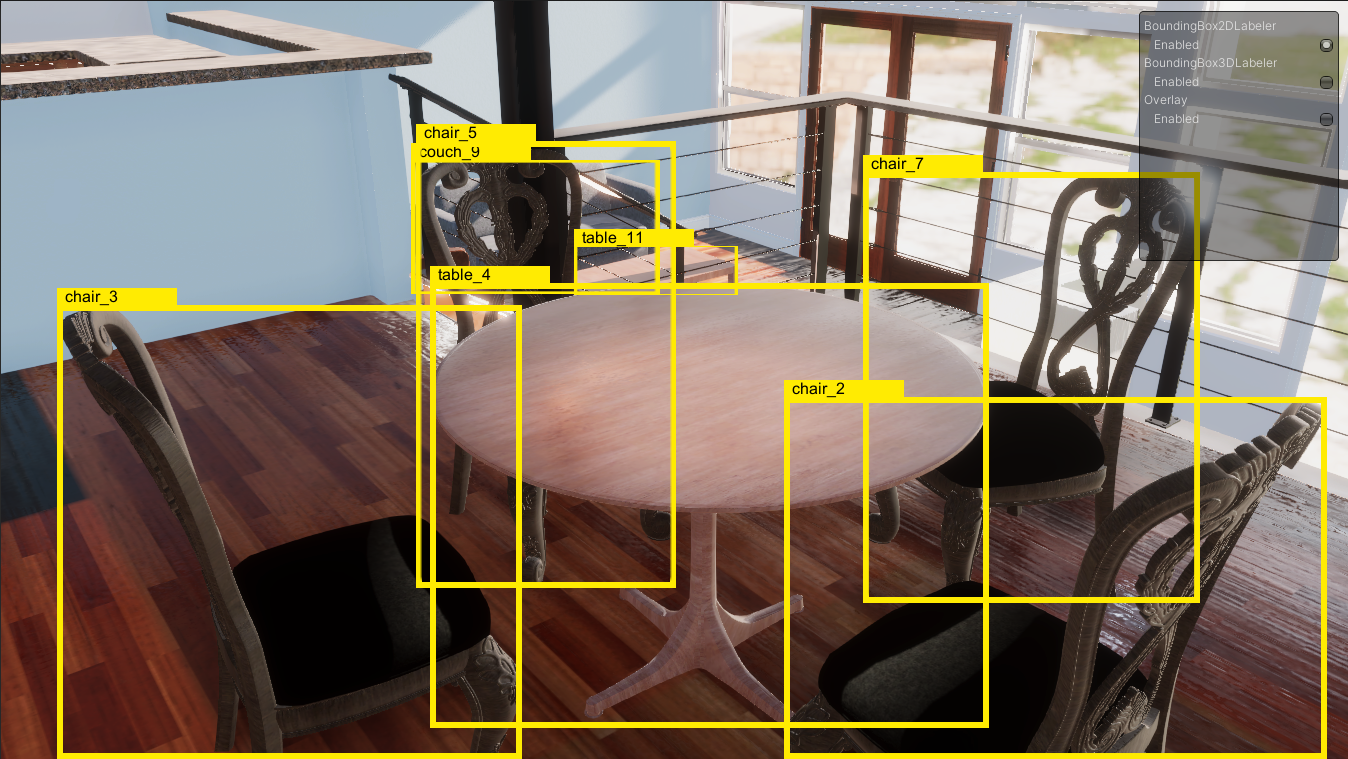} &  \includegraphics[width=38mm]{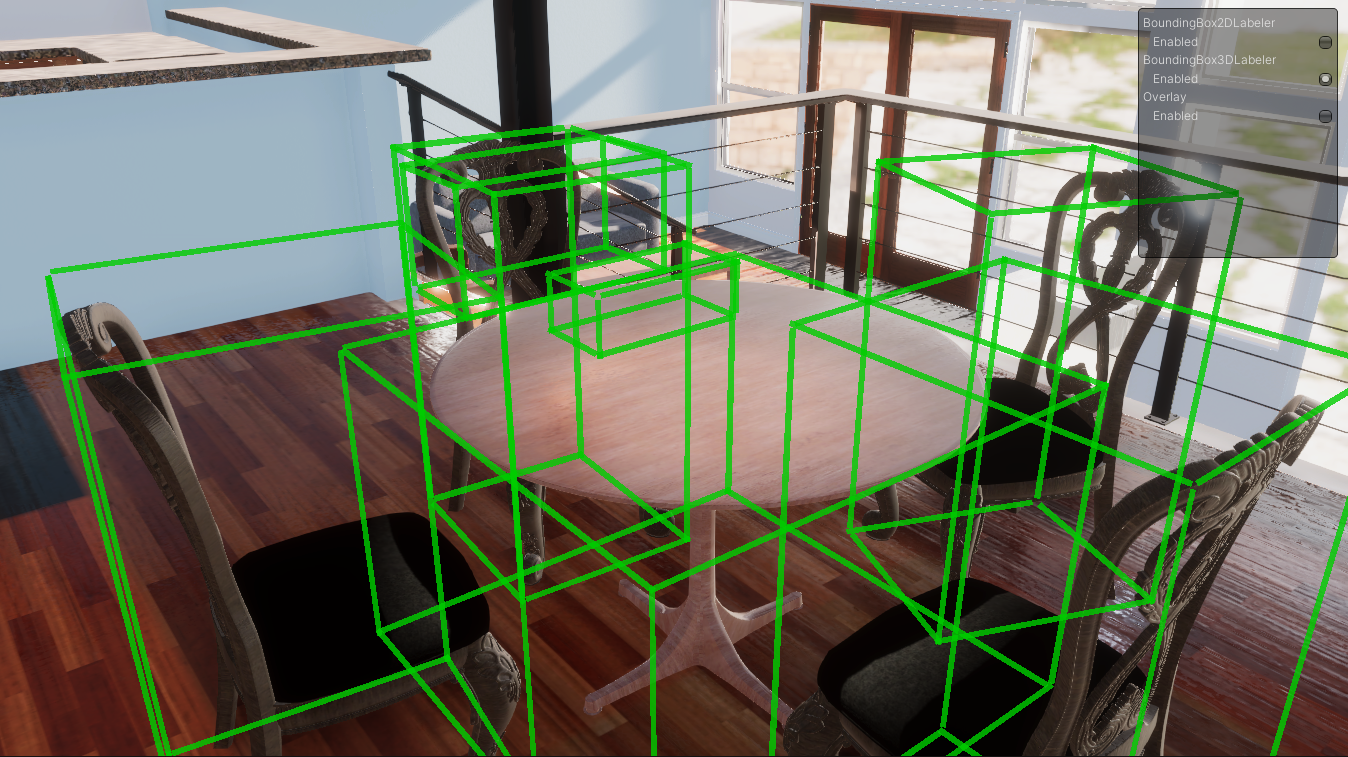} \\
 \vspace{1em}
 (a) 2D Bounding Boxes & (b) 3D Bounding Boxes \\
 \includegraphics[width=38mm]{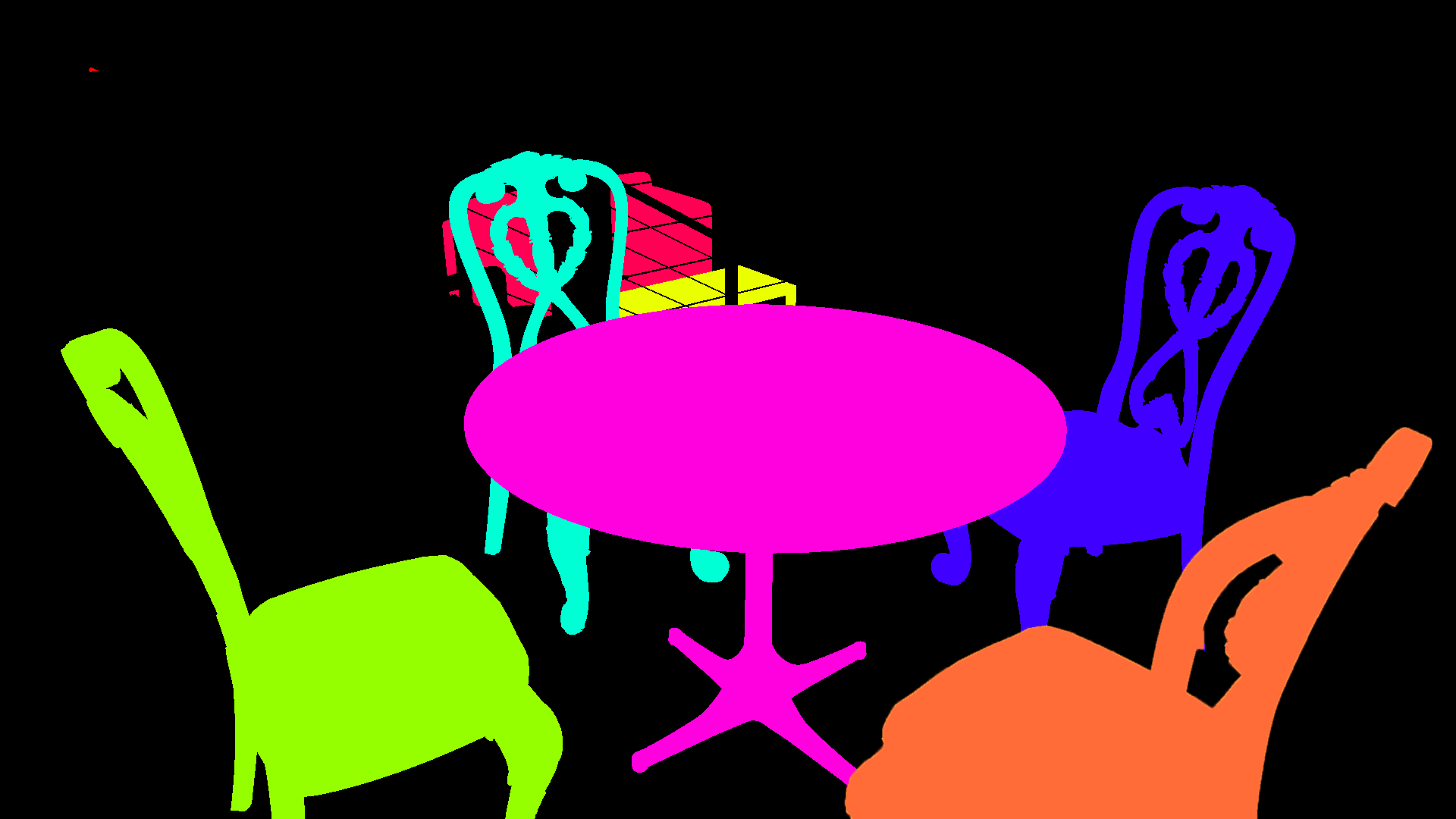} &   \includegraphics[width=38mm]{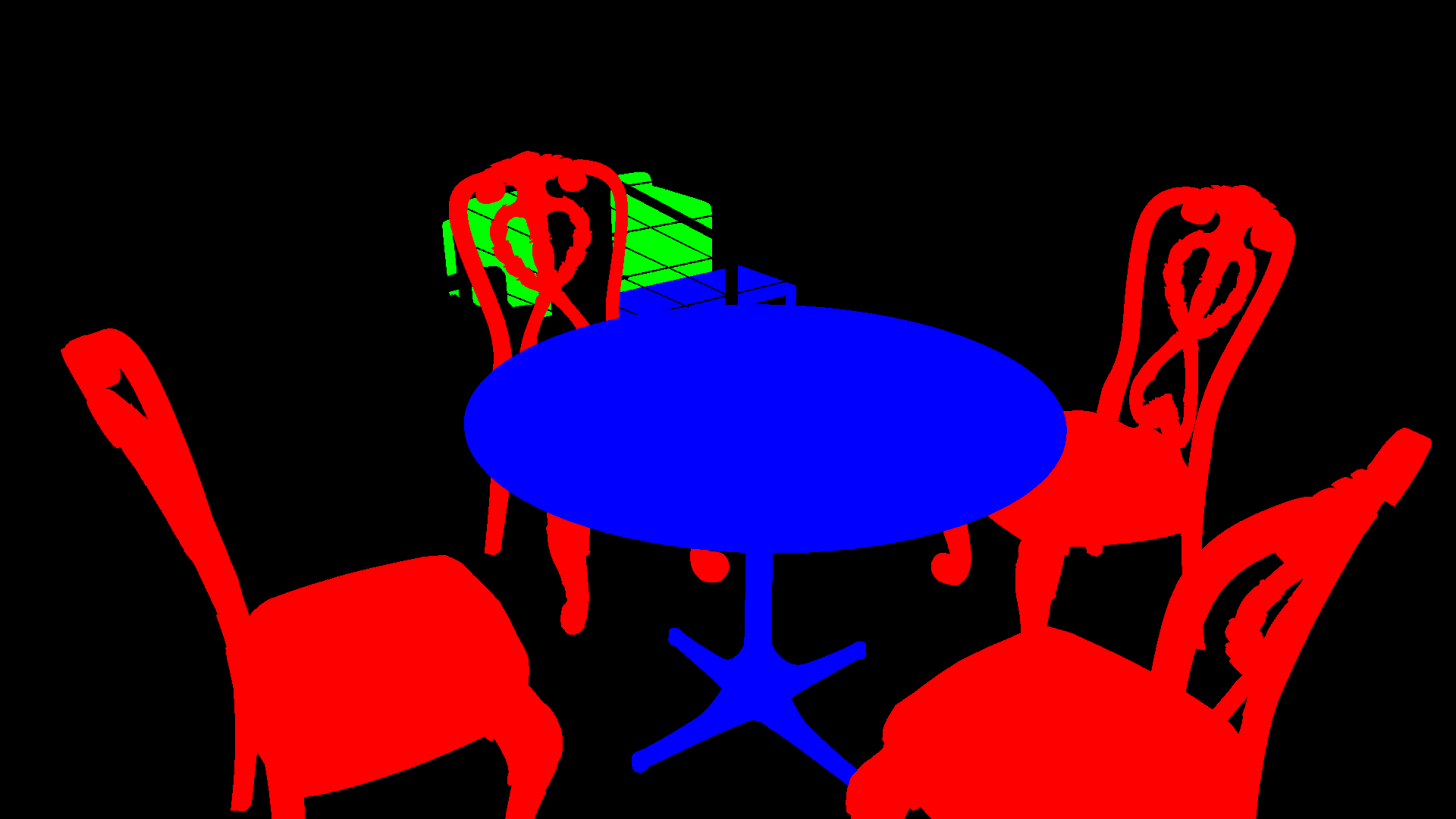} \\
 \vspace{1em}
 (c) Instance Segmentation & (d) Semantic Segmentation \\
 \multicolumn{2}{c}{\includegraphics[width=40mm]{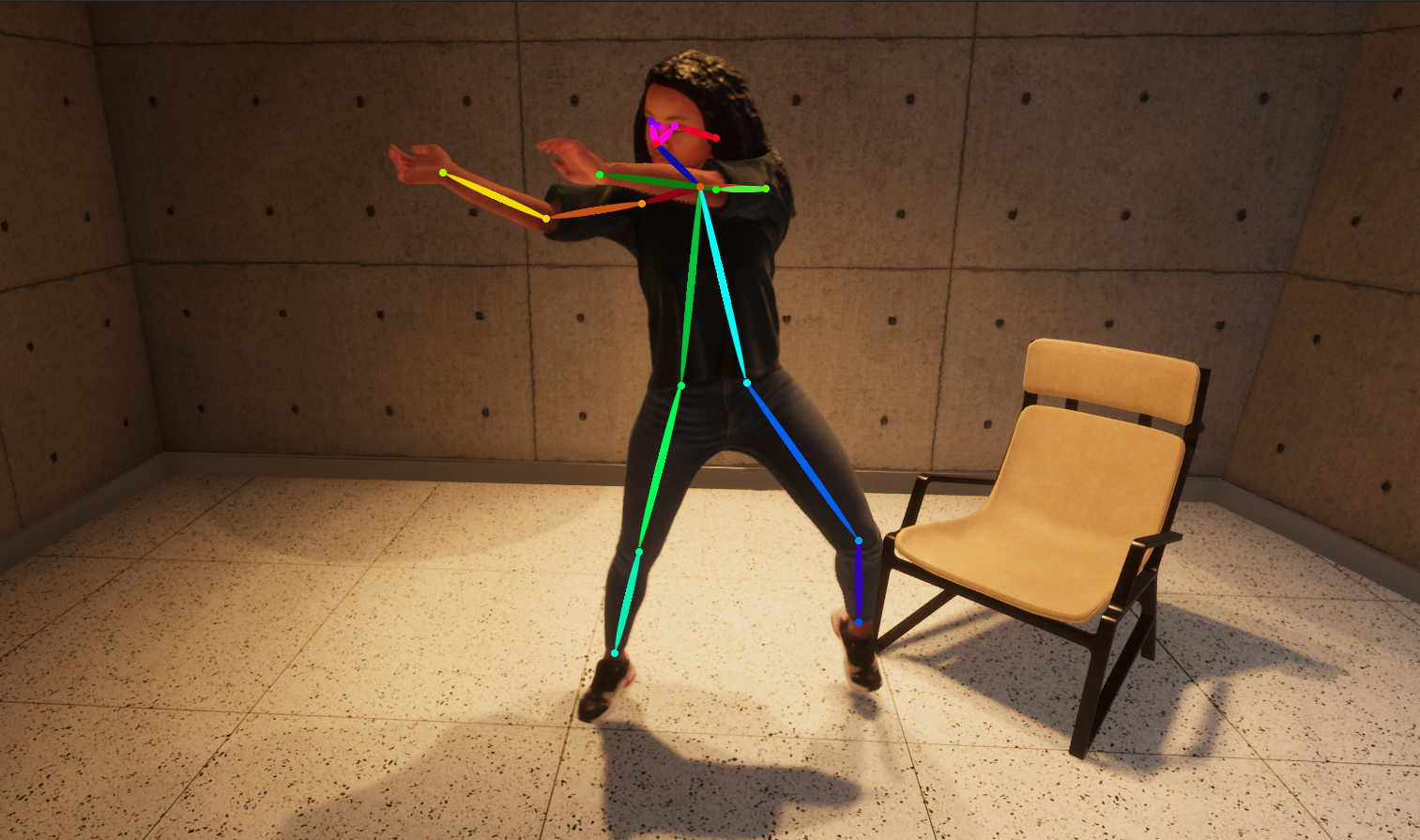} } \\
 \multicolumn{2}{c}{(e) Keypoints (points attached to objects)}  \\
\end{tabular}
\caption{Provided Labelers in the Perception package}
\label{fig:tasks}
\end{figure}

The Unity Perception package extends the Unity Editor with tools for generating synthetic datasets that include ground truth annotations. In addition, the package supports domain randomization for introducing variety into the generated datasets. Out of the box, the package supports various computer vision tasks including 2D/3D object detection, semantic segmentation, instance segmentation, and keypoints (nodes and edges attached to 3D objects, useful for tasks such as human-pose estimation). Figure \ref{fig:tasks} depicts sample outputs for these tasks. In addition, users can extended the package using \Csh{} to support new ground truth labeling methods and randomization techniques. Additionally, along with the capability to generate datasets locally, the Perception package has easy-to-use built-in support for running dataset generation jobs in the cloud using Unity Simulation, making it possible to generate millions of annotated images relatively quickly, and without the need for powerful local computing resources.

\subsection{Ground truth generation}
\label{sec:labeling}
The Perception package includes a set of Labelers which capture ground truth information along with each captured frame. The built-in Labelers support a variety of common computer vision tasks (see Table \ref{tab:tasks}). The package also includes extensible components for building new Labelers to support additional tasks using \Csh{}. Labelers derive ground truth data from labels specified on the 3D assets present in the scene. The user manually annotates the  project's library of assets with semantic labels such as ``chair'' and ``motorcycle'' to provide the Labelers with the data required to capture datasets that match the target task. The Labelers are configured with a mapping from the human readable semantic labels to the numeric canonical class ids used to train the target model. This mapping allows for labeled assets to be used across many dataset generators and labeling strategies. During simulation, these Labelers compute ground truth based on the state of the 3D scene and the rendering results, through a custom rendering pipeline. Appendix \ref{app-diagrams} provides a high-level component diagram for the ground-truth generation system.

\begin{figure*}[t]
  \begin{subfigure}{0.5\textwidth}
    \setlength{\fboxsep}{5pt}
    \setlength{\fboxrule}{1pt}
    \fbox{\includegraphics[width=0.9\linewidth]{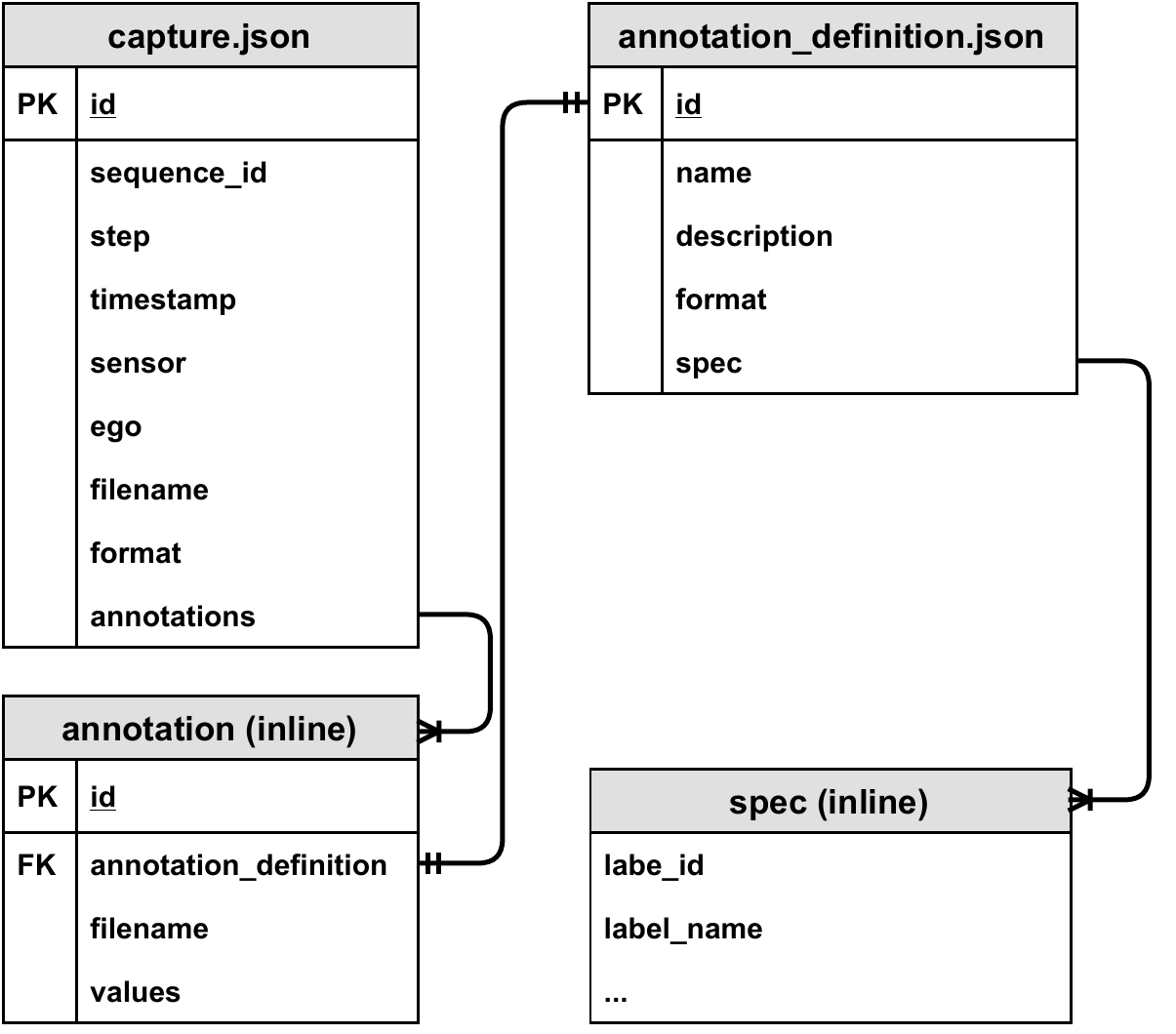}}
    \caption{Captures}
    \label{fig:cpatures}
  \end{subfigure}
  \begin{subfigure}{0.5\textwidth}
    \setlength{\fboxsep}{5pt}
    \setlength{\fboxrule}{1pt}
    \fbox{\includegraphics[width=0.9\linewidth]{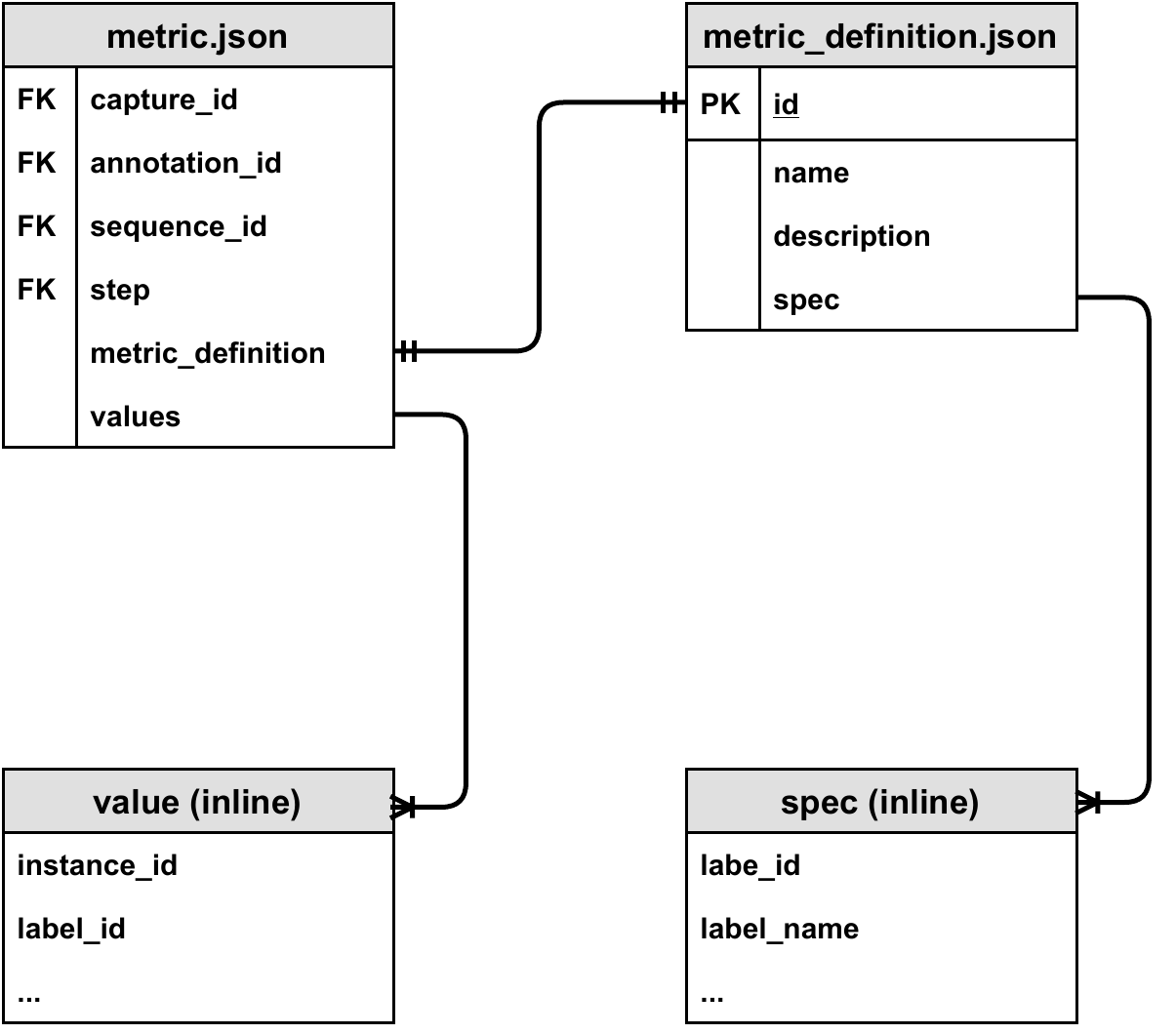}}
    \caption{Metrics}
    \label{fig:metrics}
  \end{subfigure}

  \caption{Synthetic dataset entity relation diagram for captures and metrics. (a) The captures.json files contain one to many relationships between sensor outputs and Labeler outputs (annotations). Each capture and annotation are assigned a unique identifier that allows users to look up extra data in the metrics.json files. The annotation\_definition.json file contains references to decode annotation records programmatically. Each mapping is assigned a unique identifier to distinguish between different annotation types. (b) The metrics.json files contain extra data to describe a particular capture or annotation. These extra data can be used to compute dataset statistics.}
  \label{fig:schema-diagram}
\end{figure*}

\subsection{Randomization tools}
\label{sec:randomizer}

The Unity Perception package provides a randomization framework that simplifies introducing variation into synthetic environments, leading to varied data. An entity called the \textit{Scenario} controls and coordinates all randomizations in the scene. This involves triggering a set of \textit{Randomizers} in a predetermined order. Each Scenario's execution is called an \textit{Iteration} and each Iteration can run for a user-defined number of frames. Users can configure Randomizers to act at various timestamps during each Iteration, including at the start and end or per each frame. The Randomizers expose the environment parameters for randomization and utilize samplers to pick random values for these parameters. The combination of the Scenario and its Randomizers allow the user to define elaborate and deterministic schedules for randomizations to occur throughout a simulation. 

Most pieces of the randomization framework are customizable and extensible. Users can create new Randomizers by extending the base Randomizer class to control various parameters in their environments. The Perception package comes with several sample Randomizers to assist with common randomization tasks (e.g. random object placement, position, rotation, texture, and hue), and examples on extending and customizing Randomizers. Besides, users can extend the Scenario class to include custom scheduling behavior. Users can also control the sampling strategy for each individual randomized parameter by either selecting from a set of provided distributions (including normal and uniform), or providing custom distribution curves by graphically drawing them.

The randomization framework has built-in support for distributed data generation. Users can scale randomized simulations to the cloud by launching a Unity Simulation run directly from Unity Editor. The toolset uses deterministic random sampling and ordering of randomizations to ensure that data generated during distributed cloud execution is reproducible in Unity Editor for debugging purposes.

Appendix \ref{app-diagrams} includes high-level component and sequence diagrams for Perception's randomization framework.

\subsection{Schema}

The synthetic datasets generated using the Unity Perception package typically include two types of data: simulated sensor outputs and Labeler outputs. Inspired by\cite{caesar_nuscenes_2020, lin_microsoft_2014}, we used JSON files with lightweight and extensible schema design to connect sensor and Labeler output files. This lightweight design allows us to programmatically read data while maintaining flexibility to support various computer vision tasks. Figure \ref{fig:schema-diagram} shows the dataset schema generated by the Unity Perception package. The package documentation\footnote{\url{https://github.com/Unity-Technologies/com.unity.perception/blob/master/com.unity.perception/Documentation~/Schema/Synthetic_Dataset_Schema.md}} includes the complete schema design.


\begin{figure*}[t]
  \begin{subfigure}{0.32\textwidth}
    \includegraphics[width=\columnwidth]{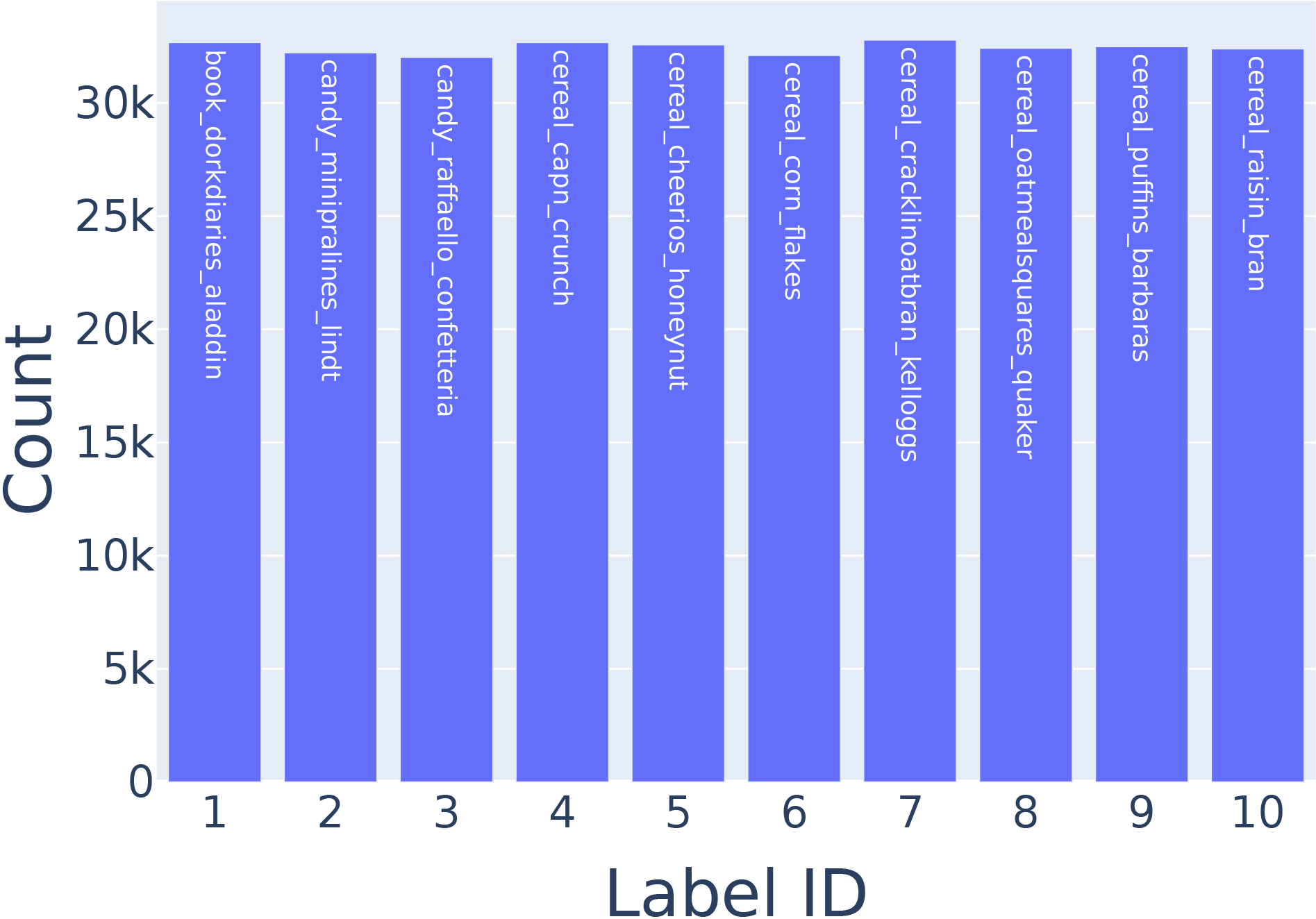}
    \caption{}
    \label{fig:total-object-counts}
  \end{subfigure}
  \hspace{5pt}
  \begin{subfigure}{0.32\textwidth}
    \includegraphics[width=\columnwidth]{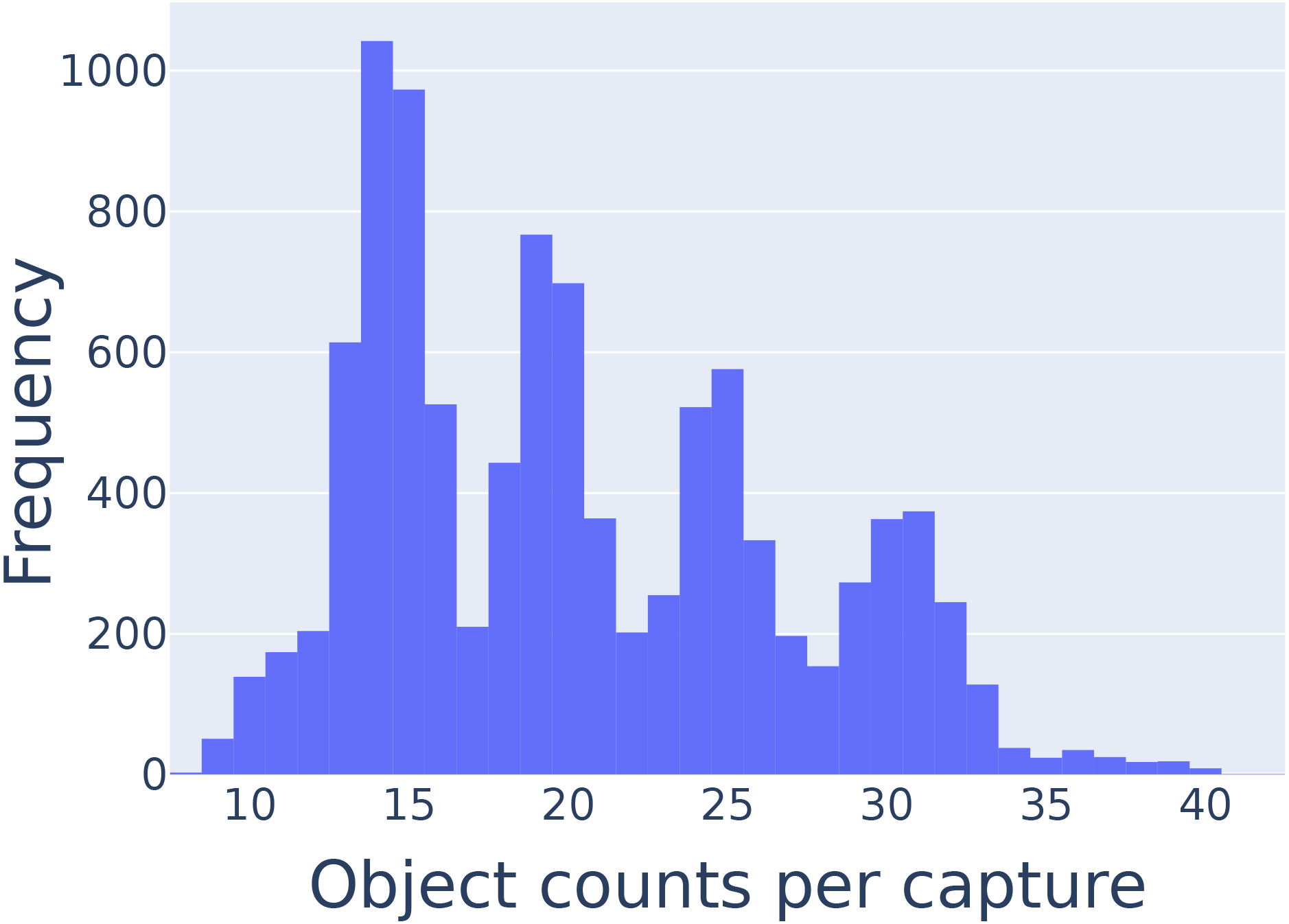} 
    \caption{}
    \label{fig:per-capture}
  \end{subfigure}
  \hspace{5pt}
  \begin{subfigure}{0.32\textwidth}
    \includegraphics[width=\columnwidth]{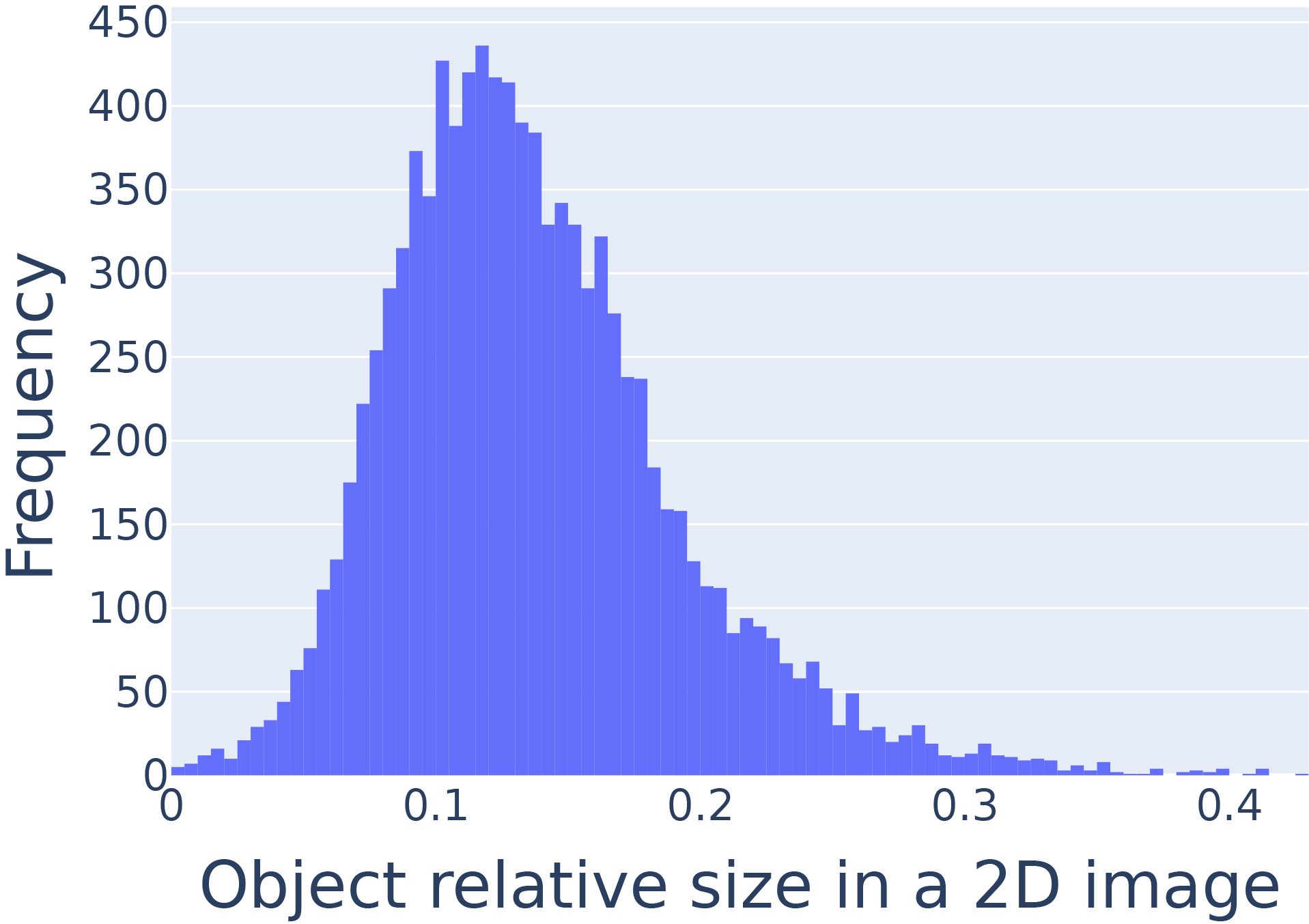}
    \caption{}
    \label{fig:object-size}
  \end{subfigure}
  
  \caption{Dataset statistics from the SynthDet example project (described in Section \ref{sec:synthdet}), providing insight into the generated dataset. (a) Total object counts aggregated per label in the dataset. The counts are nearly identical for each label since we sample objects uniformly in the generation process. The chart here depicts only the first ten labels. (b) Distribution of the number of objects per capture (label agnostic). (c) Distribution of sizes of objects relative to the 2D images. Users can control the size of objects through Randomizers. Here, $\text{relative size} = \sqrt{\frac{\text{object occupied pixels}}{\text{total image pixels}}}$.}
  \label{fig:statistics}
\end{figure*}

\subsection{Dataset Insights}
\label{sec:dataset-insights}

In addition to the perception package, we provide an accompanying python package named DatasetInsights\footnote{\url{https://github.com/Unity-Technologies/datasetinsights}} to assist the user in working with datasets generated using the Perception package. This includes generating and visualizing dataset statistics and performing model training tasks. To support both use cases, we constructed dataset IO modules that allow users to parse, load, and transform datasets in memory. 

The included statistics cover elements such as total and per frame object count, visible pixels per object, and frame by frame visualization of the captured ground truth. These, along with support for extending the toolset to include new statistics, make it possible to understand and verify the generated datasets before using them for model training. For instance, statistics can help users to decide whether a larger dataset or one with different domain randomizations applied is needed for the specific problem they are trying to solve. They also serve as a debugging tool to spot issues in the dataset. Statistics from the SynthDet project (Section \ref{sec:synthdet}) are shown in Figure \ref{fig:statistics}.


\section{Example Project - SynthDet}
\label{sec:synthdet}
\begin{figure*}[t]
  \centering
  \begin{subfigure}{0.48\textwidth}
    \includegraphics[width=\columnwidth]{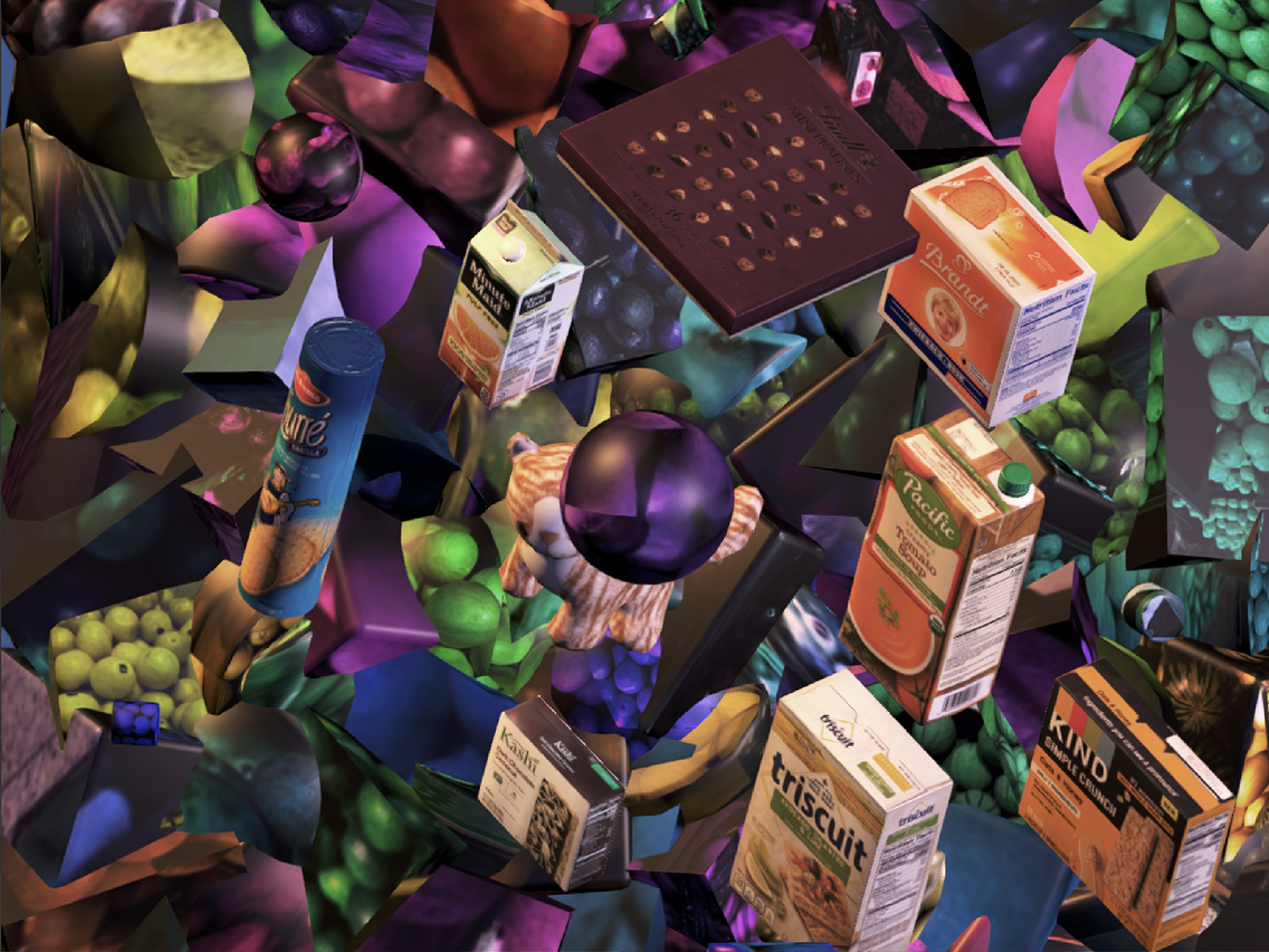}
    \caption{RGB frame output}
  \end{subfigure}
  \begin{subfigure}{0.48\textwidth}
    \includegraphics[width=\columnwidth]{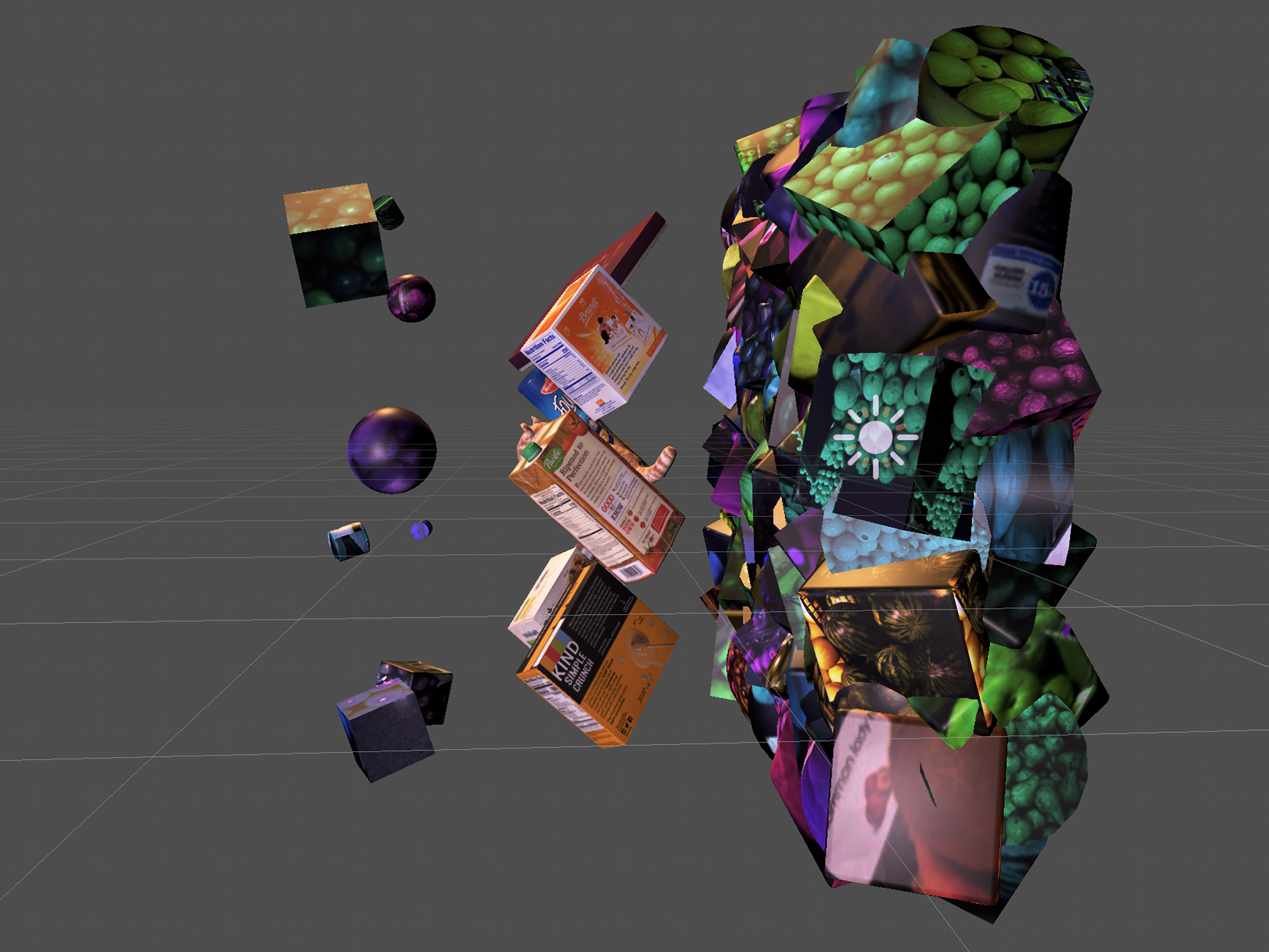}
    \caption{Side view in Unity Scene}
  \end{subfigure}
  \caption{(a) An example frame generated for the SynthDet project. The grocery products are displayed against a backdrop of objects with randomized shape, rotation, hue, and texture, and behind a foreground of occluding objects. (b) The Unity Scene for the same frame as seen from its right side. The camera is placed to the left of this view.}
  \label{fig:synthdet_example}
\end{figure*}

To prove the viability of Unity's Perception package, we built SynthDet. This project involves generating synthetic training data for a set of 63 common grocery objects, training Faster R-CNN\cite{ren_faster_2017} 2D object detection models using various combinations of synthetic and real data, and comparing and analyzing the performance of said models. This approach was inspired by previous research\cite{hinterstoisser_annotation_2019,savva_habitat_2019} and entails generating large quantities of highly randomized images in which the grocery products are placed in-between two layers of distracting objects. An example frame is shown in Figure \ref{fig:synthdet_example}.

To generate synthetic datasets, we built 3D models using 3D scans of the actual grocery objects, and imported the models into Unity Editor. In addition, we created a real-world dataset using the same products by taking numerous pictures of them in various formations and locations. In the results section, we will discuss our training approach and results using these datasets.

\subsection{Randomizations}

To achieve a randomized environment with the 3D models we used a set of Randomizers, with each undertaking a specific randomization task. In summary, the following aspects were randomized:

\begin{itemize}
\item \textbf{{Grocery (foreground) objects:}} A randomly selected subset of the 63 grocery objects is instantiated and randomly positioned in front of the camera per frame. The density of these objects is also randomized such that in some frames the objects are placed much closer to each other and there are significantly more of them compared to other frames. Furthermore, the scales of these objects are randomized on each frame, and the whole group of objects is assigned a unified random rotation.
\item \textbf{Background objects:} A group of primitive 3D objects are randomly placed close to each other, creating a ``wall'' of intersecting shapes behind the grocery objects. These objects also have a random texture, chosen from a set of 530 varied images of fruits and vegetables\cite{klasson_hierarchical_2019} , applied to them on each frame. Additionally, the rotations and color hues of these objects are randomized per frame.
\item \textbf{Occluding objects:} A set of foreground occluding objects are placed randomly at a distance closer to the camera. These are the same primitive objects used for the background, but placed farther apart. The same texture, hue, and rotation randomizations applied to the background are also applied to these occluders.
\item \textbf{Lighting:} A combination of four directional lights illuminate the scene. All of the lights have randomized intensity and color, and one has randomized rotation as well. Three of the lights affects all objects, while one significantly brighter light only affects the background objects. This light is switched on with a small probability, resulting in the background becoming overexposed in some frames, leading to more visual separation between it and the grocery objects. This reflects the real test dataset in which some frames contain much less distraction that others.
\item \textbf{Camera post processing:} The contrast and saturation of the output are randomized in small percentages. Additionally, in some frames, a small amount of blur is applied to the camera to simulate real test images.
\end{itemize}

\subsection{Results}
Using the approach discussed above, we generated a randomized synthetic dataset containing 400,000 images and 2D bounding box annotations. We also collected and annotated a real-world dataset, named UnityGroceries-Real\footnote{\url{https://github.com/Unity-Technologies/SynthDet/blob/master/docs/UnityGroceriesReal.md}}, which contains 1267 images of the 63 target grocery items. This dataset is primarily used to assess the model's performance trained on different combinations of synthetic and real-world data. The dataset was annotated using VGG Image Annotator (VIA)\cite{dutta_via_2019} with the guidelines from the PASCAL VOC dataset. We randomly shuffled and split all annotated images into a training set of 760 (60\%) images, a validation set of 253 (20\%) images, and a testing set of 254 (20\%) images. We also built a few randomly selected subsets of size 76 (10\%), 380 (50\%), and 760 (100\%) from the training set for model fine-tuning using limited amounts of the training data.

\begin{table*}[hbt!]
    \centering
    {\scriptsize
    \begin{tabularx}{\textwidth}{|l|X|X|X|X|X|X|}
    \hline
    Training Data (number of training examples) & mAP(error) & $\Delta \text{mAP}$(p-value)& $\text{mAP}^{\text{IoU50}}$(error) & $\Delta \text{mAP}^{\text{IoU50}}$(p-value) & $\text{mAR}^{\text{max=100}}$(error) & $\Delta \text{mAR}^{\text{max=100}}$(p-value)\\
    \hline
    Real-World \emph{baseline} (760)      & 0.450 (0.020) & -  & 0.719 (0.020) & -          & 0.570 (0.015) & -\\
    Synthetic (400,000)                    & 0.381 (0.013) & -0.069 (2e-4) & 0.538 (0.019) & -0.182 (4e-2) & 0.487 (0.016) & -0.082 (3e-5)\\
    Synthetic (400,000) + Real-World (76)  & 0.528 (0.006) & +0.079 (3e-5) & 0.705 (0.008) & -0.014 (1e-1) & 0.636 (0.005) & +0.066 (1e-5)\\
    Synthetic (400,000) + Real-World (380) & 0.644 (0.004) & +0.194 (2e-8) & 0.815 (0.005) & +0.095 (6e-6) & 0.732 (0.004) & +0.162 (1e-8)\\
    Synthetic (400,000) + Real-World (760) & 0.684 (0.006) & +0.234 (6e-9) & 0.854 (0.007) & +0.135 (5e-7) & 0.757 (0.006) & +0.187 (4e-9)\\
    \hline
    \end{tabularx}
    }
    \caption{Detection performance ({\small mAP, $\text{mAP}^{\text{IoU50}}$, $\text{mAR}^{\text{max=100}}$}) evaluated on the testing set of the UnityGrocreies-Real dataset. The mean and standard deviation of these metrics over 5 repeated model training procedures under different mixtures of real-world and synthetic datasets are reported in this table. We also provide the mean differences between each training strategy and real-world (baseline) with the p-value from two-sided t-tests.}
    \label{tab:synthdet-result}
\end{table*}

We use the Faster R-CNN\cite{ren_faster_2017} model with the ResNet50\cite{he_deep_2016} backbone pre-trained on the ImageNet\cite{russakovsky_imagenet_2015} dataset using the pytorch/torchvision\cite{paszke_pytorch_2019} implementation. Three training strategies with different combinations of synthetic and real-world data are used in this project. In the first strategy, only the training set of the real-world dataset is used to update the model. These models are trained with a minibatch size of 8 on 1 NVIDIA-V100 GPU for 100 epochs. We selected the best model with the lowest multi-task loss\cite{ren_faster_2017} on the validation set of the real-world data. In the second strategy, we use only the 400,000 synthetic data for model training and validation. The dataset is split into 90\% for training and 10\% for validation. These models are trained with a minibatch size of 4 on 8 NVIDIA-V100 GPUs for 10 epochs. We selected the best model with the lowest multi-task loss on the validation set of the synthetic dataset. In the third strategy, we start from the model trained and selected in strategy two and fine-tune with various subsets of the real-world training set. These models are trained with a minibatch size of 8 on 1 NVIDIA-V100 GPU for 30 epochs. We selected the best model with the lowest multi-task loss on the validation set of the real-world data. We used the Adam method for optimization\cite{kingma_adam_2015}, with $\beta_1=0.9$, $\beta_2=0.999$ in all training strategies, and an initial learning rate of $2e^{-4}$ in all strategies except the fine-tuning steps in strategy three, where we used a smaller learning rate of $2e^{-5}$. We used a gradient accumulation\cite{lin_deep_2018} of size 8 in all training strategies. All training procedures were repeated 5 times with different random weight initializations.


\begin{figure*}[t]
    \centering
    \includegraphics[width=\linewidth]{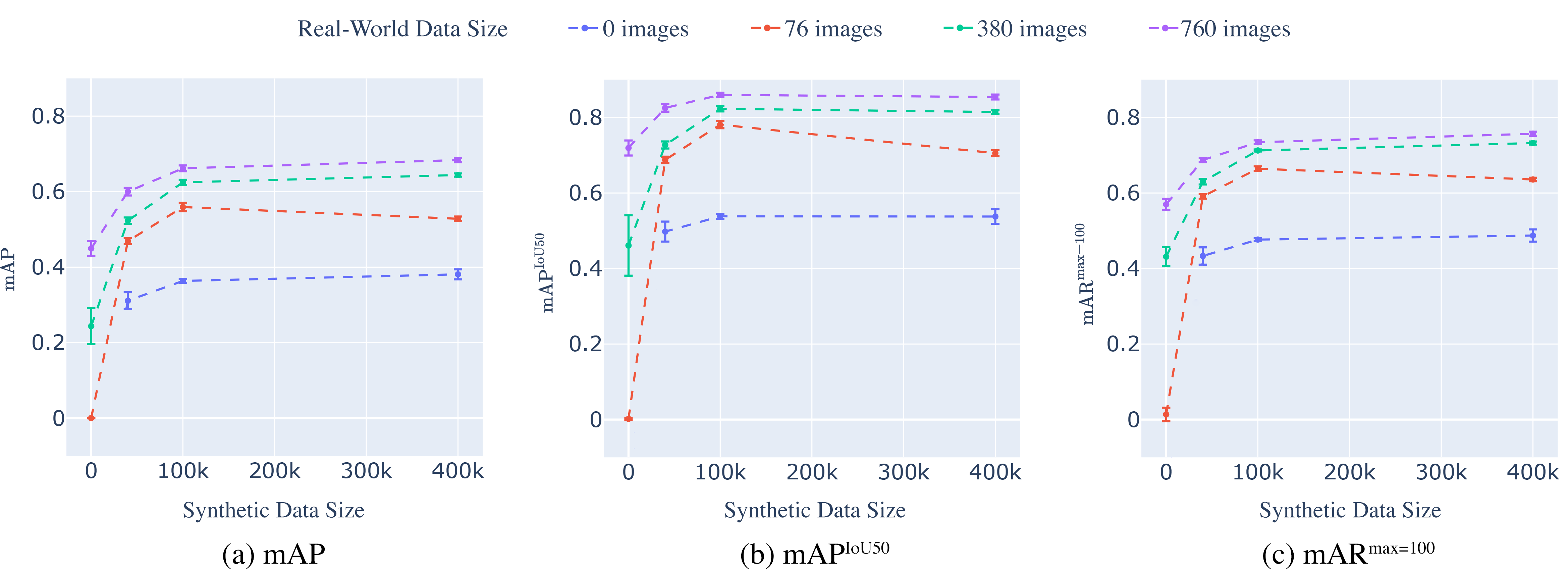}
    \caption{Detection performance ({\small mAP, $\text{mAP}^{\text{IoU50}}$, $\text{mAR}^{\text{max=100}}$}) under different combinations of synthetic (0, 40K, 100K, 400K) +  real-world (0,  76,  380,  760) data. The x-axis represents synthetic dataset size. Different colors (blue, red, green, purple) represent different sized subsets of the real-world training sets (0, 76, 380, and 760 images).}
    \label{fig:model_perf_under_diff_data}
\end{figure*}

\begin{figure*}[t]
  \includegraphics[width=0.95\linewidth]{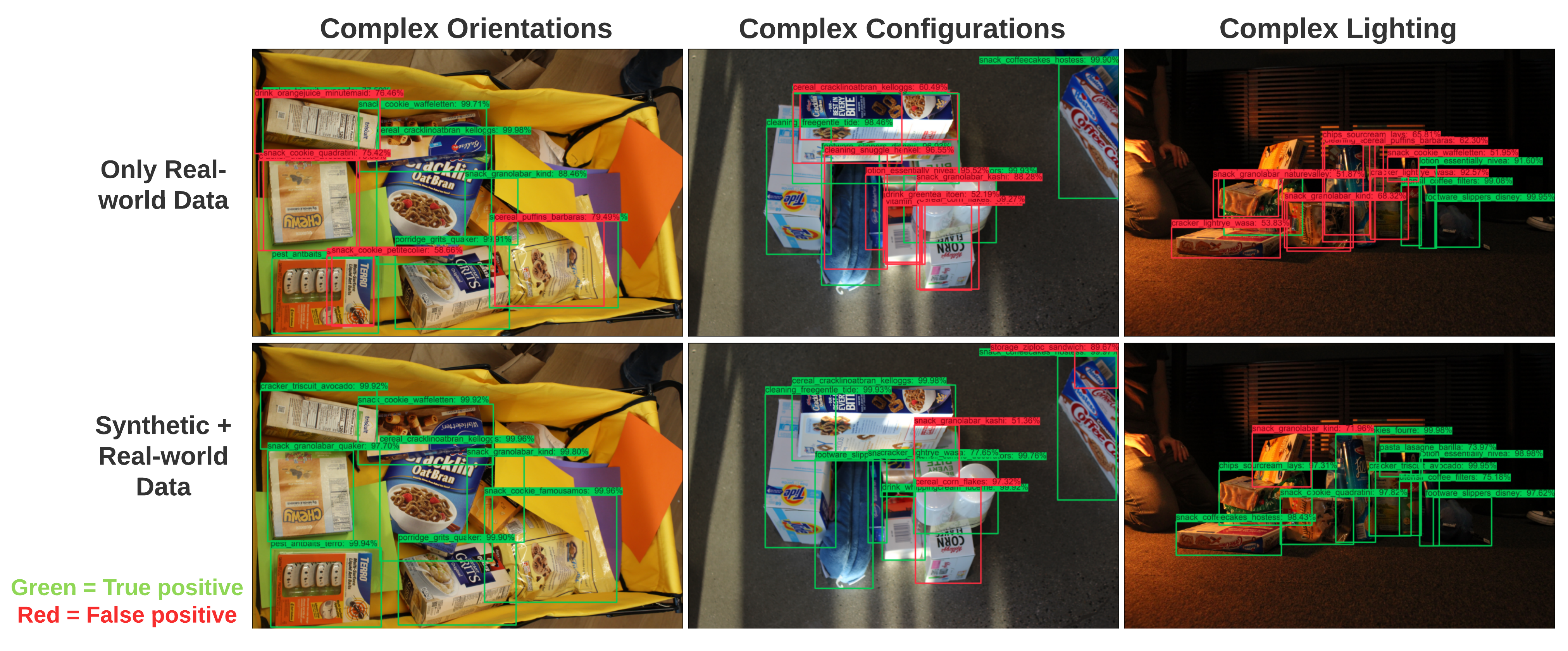}
  \caption{Illustration of model prediction quality under two different training strategies. Green bounding boxes are correct predictions, and red bounding boxes are false-positive detections. Using synthetic data improves model prediction quality in situations where the objects have complex orientations, configurations, and lighting conditions.}
  \label{fig:visualization}
\end{figure*}

Table \ref{tab:synthdet-result} presents model performance on the testing set of real-world data using different training strategies. We report three evaluation metrics: 1) mean Average Precision averaged across IoU thresholds of [0.5:0.95] ({\small mAP}), 2) mean Average Precision with a single IoU threshold of 0.5 ({\small $\text{mAP}^{\text{IoU50}}$}), and 3) mean Average Recall with a maximum detection of 100 ({\small $\text{mAR}^{\text{max=100}}$}). Figure \ref{fig:model_perf_under_diff_data} shows detection performance of models trained on different combinations of synthetic (0, 40K, 100K, 400K) + real-world (0, 76, 380, 760) data. Under each real-world data size, with more synthetic data, the model performance is improved. When using only 76 real-world images in training, the synthetic led to substantial improvements in model performance, as shown by the red lines in Figure \ref{fig:model_perf_under_diff_data}. This is while even with 760 real-world images, the addition of synthetic data helped improve model performance to a significant extent, as shown by the purple lines in Figure \ref{fig:model_perf_under_diff_data}. The model trained on 400K synthetic data and 760 real-world data showed the best performance. Figure \ref{fig:visualization} depicts some visual examples of model prediction under different training strategies. We find that the number of false-positives and false-negatives dropped significantly with synthetic + 760 real-world data. Additionally, we also see remarkable improvements in the bounding box localization. However, we still see that the model is struggling in situations with complex lighting. Detailed model performance results are provided in appendix \ref{app-performance}. Overall, our results clearly demonstrate that synthetic data can play a significant role in computer vision model training.

\section{Future Work}

There are several directions in which we would like to see this package and its ecosystem progress in order to lower the barrier of entry to computer vision research, or perhaps even enable new research avenues. 

An extensible sensor framework would make it easier to add new passive and active sensor types to support environments that rely on radar or lidar sensors, such as robotics and autonomous vehicles. Moreover, it would make it easier to implement specific well-known sensors that would simulate the behavior of their real-world counterparts. 

Continued improvement to our existing ground truth generators (Labelers) and the addition of new ones will be another area of future work. For instance, a number of actively researched computer vision tasks require data in the form of sequences of frames. An example of these is object tracking over time. Adding support for this type of output, as well as making it easy for the user to implement additional ground truth generators will be an important component of our road-map moving forward.

Research on new domains and tasks requires content which is not always readily available, as game or film assets are not necessarily readily usable in a simulation environment. We would like to formalize the definition of what it means for assets to be simulation-ready and create tools that will catalyze the process of validating assets for use in simulation. This will cover factors such as requirements on texture scales so they can be swapped at run-time, consistent pivot points which are semantically meaningful, semantic labels that can be mapped to a global taxonomy for randomized placement, and so on.

To further support researchers, we plan to improve the Linux compatibility of Unity's High Definition Render Pipeline (HDRP) and enable it to run in headless mode on Linux, the same way it currently does on Windows. The HDRP is a scriptable rendering pipeline that enables advanced global illumination algorithms that contribute to more realistic looking simulations.

Finally, we intend to make Perception simulations externally controllable from other processes potentially running on other nodes. This will allow us to support important workloads such as automated domain randomization\cite{openai_solving_2019}. We are working on a protocol as well as a way to define and run such heterogeneous distributed compute graphs in order to significantly reduce the complexity of scaling out experiments over large pools of simulation nodes.

{\small
\bibliographystyle{ieeetr}
\bibliography{perception}
}


\begin{appendices}
\section{Perception package diagrams}
\label{app-diagrams}
\begin{figure*}[h]
    \centering
    \includegraphics[width=0.85\linewidth]{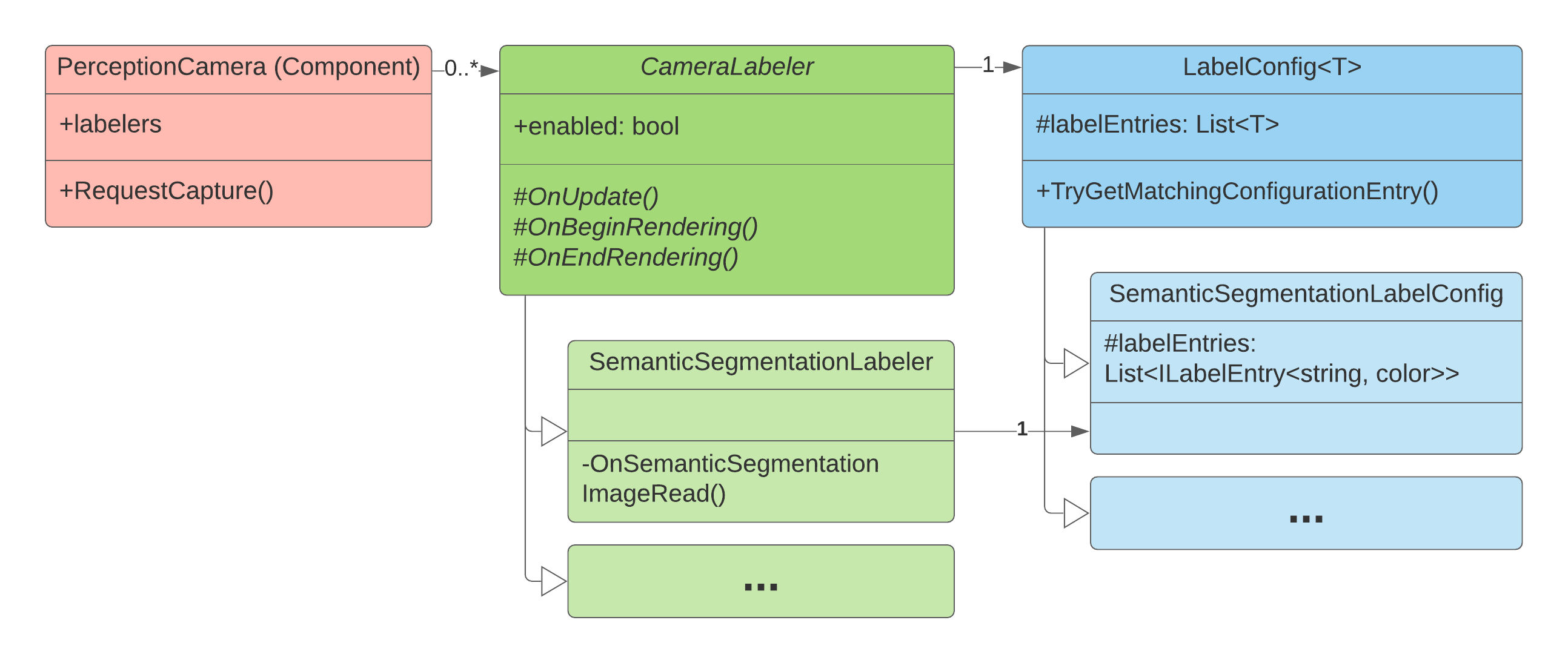}
    \caption{Class diagram for the ground truth generation system of the Perception package. A set of Camera Labelers are added to the Perception Camera, each tasked with generating a specific type of ground truth. For instance, the Semantic Segmentation Labeler outputs segmentation images in which each labeled object is rendered in a unique user-definable color and non-labeled objects and the background are rendered black. The LabelConfig acts as a mapping between string labels and object classes (currently colors or integers), deciding which labels in the scene (and thus which objects) should be tracked by the Labeler, and what color (or integer id) they should have in the captured frames. The Perception package currently comes with Labelers for five computer vision tasks (Figure \ref{fig:tasks}), and the user can implement more by extending the CameraLabeler class.}
    \label{fig:labeling_uml}
\end{figure*}

\begin{figure*}[h]
    \centering
    \includegraphics[width=\linewidth]{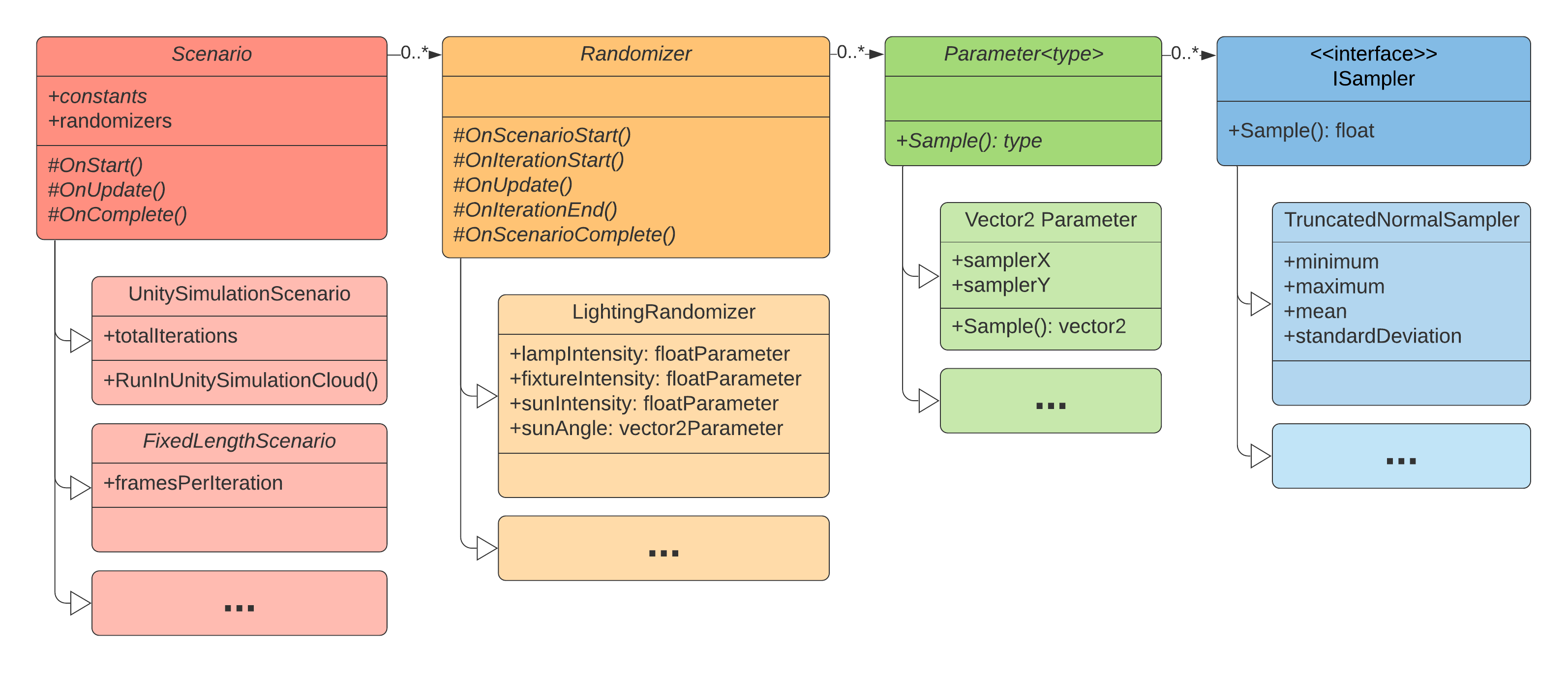}
    \caption{Class diagram for the randomization framework included in the Perception package. The Scenario coordinates the life-cycle of the simulation, executing a number of randomization Iterations. In each iteration, a group of Randomizers are triggered, each of which is tasked with randomizing one or more aspects of the simulation and the objects present in the scene. To carry out this randomization, Randomizers can include one or more Parameter objects which internally use Samplers in order to generate random typed values. The Scenario, Randomizer, and Parameter classes are all extensible. Additionally, the user can implement the ISampler interface to achieve further customization in sampling behavior if the provided distributions are not sufficient.}
    \label{fig:randomization_uml}
\end{figure*}

\begin{figure*}[h]
    \centering
    \includegraphics[width=0.37\linewidth]{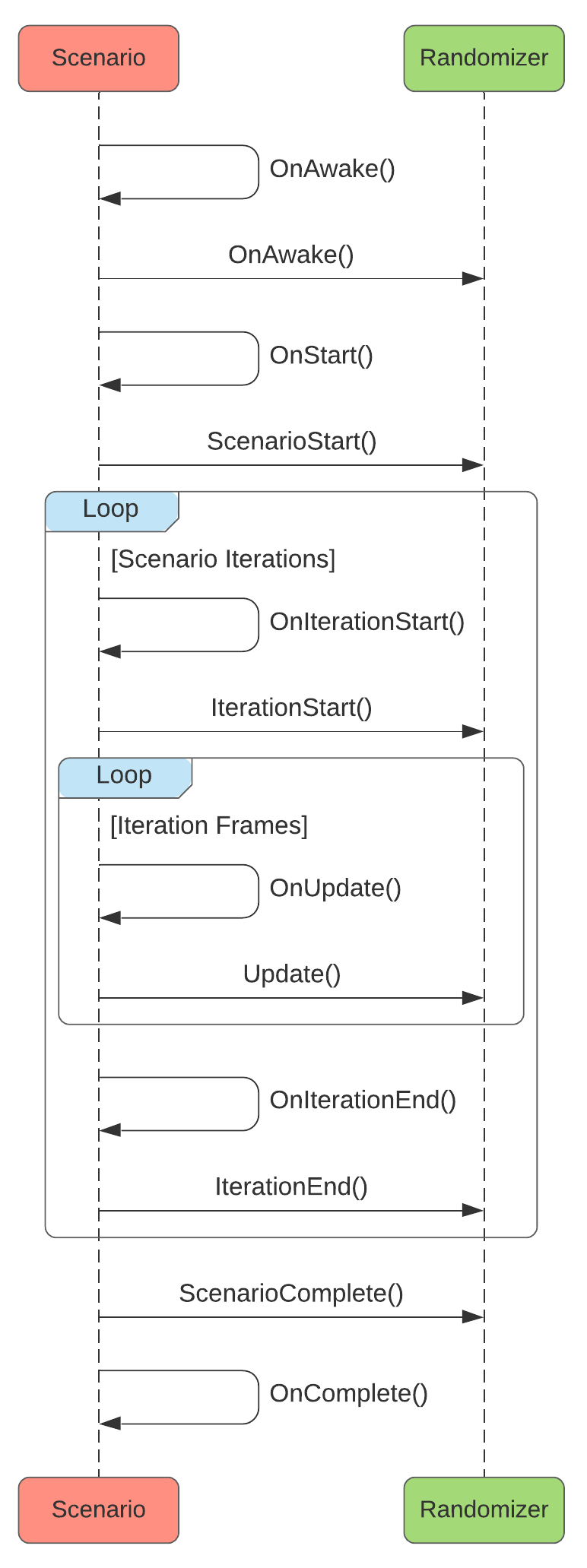}
    \caption{Sequence diagram for the execution of a Scenario, depicting the various life-cycle events that are triggered as the Scenario progresses through its Iterations, calling on its Randomizers to perform their duties. At each point in the life-cycle, the Randomizers included in a Scenario are triggered in the order they are added to the Scenario by the user. All life-cycle events in the diagram above can be overridden in order to customize the behavior of the Scenario and its Randomizers. For simplicity, the diagram above portrays a Scenario with just one Randomizer.}
    \label{fig:scenario_seq_diag}
\end{figure*}

Figures \ref{fig:labeling_uml}, \ref{fig:randomization_uml}, and \ref{fig:scenario_seq_diag} depict high-level component diagrams and a sequence diagram for the ground-truth generation and randomization systems within the Perception package. The level of detail in these diagrams are abstracted to a certain extent, to make them easier to digest here.

\section{SynthDet model performance}
\label{app-performance}
Table \ref{tab:synthdet-result} provides detailed performance figures for the SynthDet model under a variety of combinations of real and synthetic training data.

\begin{table*}[hbt!]
    \centering
    {\scriptsize
    \begin{tabularx}{\textwidth}{|l|X|X|X|X|X|X|}
    \hline
    Training Data (number of training examples) & mAP(error) & $\Delta\text{mAP}$(p-value)& $\text{mAP}^{\text{IoU50}}$(error) & $\Delta\text{mAP}^{\text{IoU50}}$(p-value) & $\text{mAR}^{\text{max=100}}$(error) & $\Delta\text{mAR}^{\text{max=100}}$(p-value)\\
    \hline
    Real-World \emph{baseline} (760)      & 0.450 (0.020) & -  & 0.719 (0.020) & -          & 0.570 (0.015) & -\\
    Real-World (76)                       & 0.001 (0.001)  &  -0.449 (3e-11)  &  0.002 (0.003)  &  -0.717 (7e-13)  &  0.014 (0.018)  &  -0.556 (2e-11) \\
    Real-World (380)  &  0.244 (0.048)  &  -0.206 (2e-5)  &  0.461 (0.080)  &  -0.258 (1e-4)  &  0.431 (0.025)  &  -0.138 (5e-6) \\
    Synthetic (40,000)  &  0.311 (0.023)  &  -0.139 (7e-6)  &  0.498 (0.026)  &  -0.221 (4e-7)  &  0.433 (0.023)  &  -0.137 (4e-6) \\
    Synthetic (100,000)  &  0.364 (0.005)  &  -0.086 (1e-5)  &  0.538 (0.007)  &  -0.181 (5e-8)  &  0.477 (0.004)  &  -0.093 (7e-7) \\
    Synthetic (400,000)  &  0.381 (0.013)  &  -0.069 (2e-4)  &  0.538 (0.019)  &  -0.182 (5e-7)  &  0.487 (0.004)  &  -0.082 (3e-5) \\
    Synthetic (40,000) + Real-World (76)  &  0.469 (0.008)  &  0.019 (7e-2)  &  0.688 (0.008)  &  -0.31 (1e-2)  &  0.591 (0.006)  &  0.021 (2e-2) \\
    Synthetic (100,000) + Real-World (76)  &  0.523 (0.008)  &  0.073 (6e-5)  &  0.727 (0.009)  &  0.008 (4e-1)  &  0.630 (0.007)  &  0.060 (3e-5) \\
    Synthetic (400,000) + Real-World (76)  &  0.528 (0.006)  &  0.078 (2e-8)  &  0.705 (0.002)  &  -0.014 (1e-1)  &  0.636 (0.004)  &  0.066 (1e-5) \\
    Synthetic (40,000) + Real-World (380)  &  0.559 (0.011)  &  0.109 (5e-6)  &  0.781 (0.009)  &  0.062 (2e-4)  &  0.664 (0.006)  &  0.095 (9e-7) \\
    Synthetic (100,000) + Real-World (380)  &  0.625 (0.048)  &  0.175 (6e-8)  &  0.823 (0.007)  &  0.104 (4e-6)  &  0.713 (0.003)  &  0.143 (2e-8) \\
    Synthetic (400,000) + Real-World (380)  &  0.644 (0.004)  &  0.194 (2e-8)  &  0.815 (0.005)  &  0.095 (6e-6)  &  0.732 (0.004)  &  0.162 (1e-8) \\
    Synthetic (40,000) + Real-World (760)  &  0.600 (0.010)  &  0.150 (4e-7)  &  0.825 (0.010)  &  0.106 (5e-6)  &  0.687 (0.006)  &  0.118 (1e-7) \\
    Synthetic (100,000) + Real-World (760)  &  0.662 (0.008)  &  0.212 (2e-8)  &  0.860 (0.006)  &  0.141 (3e-7)  &  0.734 (0.005)  &  0.164 (1e-8) \\
    Synthetic (400,000) + Real-World (760)  &  0.684 (0.006)  &  0.234 (6e-9)  &  0.854 (0.007)  &  0.135 (5e-7)  &  0.757 (0.005)  &  0.187 (4e-9) \\
    \hline
    \end{tabularx}
    }
    \caption{Detection performance ({\small mAP, $\text{mAP}^{\text{IoU50}}$, $\text{mAR}^{\text{max=100}}$}) evaluated on the testing set of the UnityGrocreies-Real dataset. This table reports the mean and standard deviation of all metrics over 5 repeated model training procedures, for each of the included combinations of real and synthetic data.}
    \label{tab:synthdet-result}
\end{table*}




\end{appendices}

\end{document}